\documentclass[preprint,12pt]{elsarticle}

\usepackage{amssymb}
\usepackage{tabularx}
\usepackage{booktabs}
\usepackage{subfigure}
\usepackage{amsmath}
\usepackage{epsfig}
\usepackage{longtable}
\usepackage{lscape}
\usepackage{epsf}
\usepackage{multirow}
\usepackage{slashbox}
\usepackage{array}
\usepackage{dcolumn,longtable,hhline,colortbl}
\usepackage[table]{xcolor}
\usepackage[rotateright]{rotating}
\usepackage[section] {placeins}
\usepackage{morefloats}

\newdefinition{definition}{Definition}
\newdefinition{example}{Example}
\newdefinition{property}{Property}

\journal{Expert Systems with Applications}

\begin{document}

\begin{frontmatter}



\title{D-CFPR: D numbers extended consistent fuzzy preference relations}

\author[SWU]{Xinyang Deng}
\author[HKPU]{Felix T.S. Chan}
\author[UBC]{Rehan Sadiq}
\author[VU]{Sankaran Mahadevan}
\author[SWU,VU]{Yong Deng\corref{COR}}
\ead{ydeng@swu.edu.cn; prof.deng@hotmail.com}
\cortext[COR]{Corresponding author: Yong Deng, School of Computer and Information Science, Southwest University, Chongqing 400715, China.}
\address[SWU]{School of Computer and Information Science, Southwest University, Chongqing, 400715, China}
\address[HKPU]{Department of Industrial and Systems Engineering, The Hong Kong Polytechnic University, Hong Kong, China}
\address[UBC]{School of Engineering, University of British Columbia, 1137 Alumni Ave, Kelowna, BC V1V 1V7, Canada}
\address[VU]{School of Engineering, Vanderbilt University, Nashville, TN, 37235, USA}

\begin{abstract}
How to express an expert's or a decision maker's preference for alternatives is an open issue. Consistent fuzzy preference relation (CFPR) is with big advantages to handle this problem due to it can be construed via a smaller number of pairwise comparisons and satisfies additive transitivity property. However, the CFPR is incapable of dealing with the cases involving uncertain and incomplete information. In this paper, a D numbers extended consistent fuzzy preference relation (D-CFPR) is proposed to overcome the weakness. The D-CFPR extends the classical CFPR by using a new model of expressing uncertain information called D numbers. The D-CFPR inherits the merits of classical CFPR and can be totally reduced to the classical CFPR. This study can be integrated into our previous study about D-AHP (D numbers extended AHP) model to provide a systematic solution for multi-criteria decision making (MCDM).

\end{abstract}

\end{frontmatter}

\section{Introduction}
Preference relation has played a fundamental role in most decision processes \cite{Xu2007surverypr}. According to previous studies, the preference relation can be divided into two categories. The first one is multiplicative preference relation \cite{herrera2001multiperson,chiclana2004induced,liu2012goal,xia2014multiplicative} which is subjected to the multiplicative reciprocal, i.e. $a_{ij} \times a_{ji} = 1$. The second one is fuzzy preference relation \cite{Tanino1984fuzzypref,gong2008least,parreiras2012dynamic,rezaei2013multi}which is described by fuzzy pairwise comparison with an additive reciprocal, i.e. $r_{ij} + r_{ji} = 1$. As the basic element of many decision making methods especially in analytic hierarchy process (AHP) model \cite{saaty1980analytic,chan2007global,ishizaka2011selection,ishizaka2013calibrated}, the preference relation has attracted many interests \cite{fernandez2010solving,evren2011multi,xu2012error,liu2012least,xu2013distance,xia2013preference,Wang2014169}.

Fuzzy preference relations \cite{herrera2007consensus,liu2012new} provide a method to construct the decision matrices of pairwise comparisons based on the linguistic values given by experts. The value given by the experts represents the degree of the preference for the first alternative over the second alternative. Assume there are $n$ alternatives, a total of $n(n-1)/2$ pairwise comparisons need to be answered for constructing a fuzzy preference relation. What's more, there always exists a potential risk that the constructed fuzzy preference relation is inconsistent due to the inability of human beings to deal with overcomplicated objects \cite{chiclana2002note,xu2013ordinal,xia2013algorithms}. In order to overcome the deficiencies, Herrera-Viedma et al. \cite{HerreraViedma2004someiss} proposed consistent fuzzy preference relation (CFPR) to construct the pairwise comparison decision matrices based on additive transitivity property \cite{Tanino1984fuzzypref,tanino1988fuzzy}, i.e. $r_{ij} + r_{jk} + r_{ki} = 3/2$. The merit of CFPR consists of two aspects. Firstly, only is a total of $n-1$ pairwise comparisons needed to construct a CFPR. Secondly, it is always consistent in a CFPR. Due to these merits, the CFPR is widely used in many fields \cite{wang2007applying,chen2012supplier,lan2012deriving}.

Although the CFPR is so advantageous to express experts' or decision makers' preferences, however, the original CFPR is constructed on the foundation of complete and certain information. It is unable to deal with the cases involving incomplete and uncertain information. For example, an expert gives that the first alternative $A_1$ is more important than the second alternative $A_2$. How to express the linguistic variable ``more important"? Some one would say it means $r_{12} = 0.7$, some others will say it seems that $r_{12} = 0.8$. Or it is more reasonable that $r_{12} = 0.7$ with a degree $x$ and $r_{12} = 0.8$ with a degree $1-x$, where $x \in [0,1]$. Besides, incomplete information means the preference values are not complete due to the lack of knowledge and the limitation of cognition. For example, an expert gives $r_{12} = 0.8$ with a belief of 0.7, the remainder 0.3 belief can not be assigned to any linguistic values due to the lack of knowledge. With respect to these cases involving uncertain and incomplete entries, the conventional CFPR is incapable.

To overcome these weaknesses, D numbers \cite{Deng2012DNumbers,Deng2014DAHPSupplier,Deng2014EnvironmentDNs,Deng2014BridgeDNs}, a new model of expressing uncertain information, has been employed in the construction of CFPR. A new preference relation called D numbers extended consistent fuzzy preference relation (short for D-CFPR) is proposed in this paper. The D-CFPR uses D numbers to express the linguistic preference values given by experts or decision makers, and it can be reduced to classical CFPR. In contrast with the construction of CFPR, a method is proposed for constructing the D-CFPR, and the proposed method is also effective to construct the CFPR. In addition, the priority weights and ranking of alternatives can be obtained from a D-CFPR based on our previous work \cite{Deng2014DAHPSupplier}. In \cite{Deng2014DAHPSupplier}, we proposed a D-AHP (D numbers extended AHP) model for multi-criteria decision making (MCDM). The work in this paper provides a solution of constructing the preference matrix for D-AHP. Based on these two studies, we systematically provide a novel solution for MCDM problems.

The rest of this paper is organized as follows. Section \ref{SectCFPR} gives a brief introduction about the consistent fuzzy preference relation. In Section \ref{SectD-CFPR}, the proposed D numbers extended consistent fuzzy preference relation is presented. Some numerical examples are given in Section \ref{SectNumeExams}. Finally, concluding remarks are given in Section \ref{SectConclusions}.

\section{Consistent fuzzy preference relations (CFPR)} \label{SectCFPR}
Fuzzy preference relations \cite{Xu2007surverypr,Tanino1984fuzzypref,herrera2007consensus,HerreraViedma2004someiss} enable an expert or a decision maker to give linguistic values for the comparison of alternatives or creteria. The preference values employed in a fuzzy preference relation are real numbers belonging to $[0, 1]$. A reciprocal fuzzy preference relation $R$ on a set of alternatives $A = \{ A_1 ,A_2 , \cdots ,A_n \}$ is represented by a fuzzy set on the product set $A \times A$, and is characterized by a membership function \cite{Xu2007surverypr,herrera2007consensus,HerreraViedma2004someiss}
\begin{equation}
\mu _R :A \times A \to [0,1].
\end{equation}
when the cardinality of $A$ is small, the preference relation may be conveniently represented by an $n \times n$ matrix $R = [r_{ij}]_{n \times n}$, being $r_{ij} = \mu_R (A_i, A_j) \forall i,j \in \{ 1,2, \cdots ,n\}$, namely
\begin{equation}\label{fuzzymatrix}
R = \begin{array}{*{20}c}
   {} & {\begin{array}{*{20}c}
   {A_1 } & {A_2 } &  \cdots  & {A_n }  \\
\end{array}}  \\
   {\begin{array}{*{20}c}
   {A_1 }  \\
   {A_2 }  \\
    \vdots   \\
   {A_n }  \\
\end{array}} & {\left[ {\begin{array}{*{20}c}
   {r_{11} } & {r_{12} } &  \cdots  & {r_{1n} }  \\
   {r_{21} } & {r_{22} } &  \cdots  & {r_{2n} }  \\
    \vdots  &  \vdots  &  \ddots  &  \vdots   \\
   {r_{n1} } & {r_{n2} } &  \cdots  & {r_{nn} }  \\
\end{array}} \right]}  \\
\end{array}
\end{equation}
where (1) $r_{ij} \ge 0$; (2) $r_{ij} + r_{ji} = 1, \; \forall i,j \in \{ 1,2, \cdots ,n\}$; (3) $r_{ii} = 0.5,\; \forall i \in \{ 1,2, \cdots ,n\}$. $r_{ij}$ denotes the preference degree of alternative $A_i$ over alternative $A_j$.
\begin{equation*}
r_{ij}  = \mu _R (A_i ,A_j ) = \left\{ {\begin{array}{*{20}l}
   0 \hfill & \text{$A_j$ is absolutely preferred to $A_i$;} \hfill  \\
   { \in (0,0.5)} \hfill & \text{$A_j$ is preferred to $A_i$ to some degree;} \hfill  \\
   {0.5} \hfill & \text{indifference between $A_i$ and $A_j$;}  \hfill  \\
   { \in (0.5,1)} \hfill & \text{$A_i$ is preferred to $A_j$ to some degree;} \hfill  \\
   1 \hfill & \text{$A_i$ is absolutely preferred to $A_j$.} \hfill  \\
\end{array}} \right.
\end{equation*}

Herrera-Viedma et al. \cite{HerreraViedma2004someiss} proposed the consistent fuzzy preference relation (CFPR) for the construction of pairwise comparison decision matrices based on additive transitivity property \cite{Tanino1984fuzzypref,tanino1988fuzzy}. A reciprocal fuzzy preference relation $R = [r_{ij}]_{n \times n}$ is called a consistent fuzzy preference relation if and only if $r_{ij} + r_{jk} + r_{ki} = 3/2, \quad \forall i < j < k$. For a CFPR $R = [r_{ij}]_{n \times n}$, the following two equations are satisfied \cite{HerreraViedma2004someiss}:
\begin{equation}\label{CFPREQ1}
r_{ij}  + r_{jk}  + r_{ki}  = \frac{3}{2},\quad \forall i < j < k.
\end{equation}
\begin{equation}\label{CFPREQ2}
r_{i(i + 1)}  + r_{(i + 1)(i + 2)}  +  \cdots  + r_{(j - 1)j}  + r_{ji}  = \frac{{j - i + 1}}{2},\quad \forall i < j.
\end{equation}

The biggest advantage of CFPRs is that it is reciprocal and consistent. Based on the results presented in Eq.(\ref{CFPREQ1}) and Eq.(\ref{CFPREQ2}), a CFPR can be constructed from the set of $n-1$ values $\{r_{12}, r_{23}, \cdots, r_{(n-1)n}\}$. That means only $n-1$ pairwise comparisons are required in the process of constructing a CFPR. Compared with the construction of the ordinary fuzzy preference relation which requires $n(n-1)/2$ pairwise comparisons, the rest of $(n-1)(n-2)/2$ pairwise comparisons are computed by using additive transitivity property in the construction of CFPRs. It is noted that the values in the generated CFPR may do not fall in the interval $[0, 1]$, but fall in an interval $[-a, 1+a]$, $a>0$. In such a case, the values in the obtained CFPR need to be transformed by using a transformation function that preserves reciprocity and additive consistency. The transformation function is defined as \cite{HerreraViedma2004someiss}:
\begin{equation}\label{Transfunction}
f:[ - a,1 + a] \to [0,1],\quad f(r) = \frac{{r + a}}{{1 + 2a}}.
\end{equation}

\section{Proposed D numbers extended consistent fuzzy preference relations (D-CFPR)}\label{SectD-CFPR}
\subsection{D numbers}
D number \cite{Deng2012DNumbers,Deng2014DAHPSupplier,Deng2014EnvironmentDNs,Deng2014BridgeDNs} is a new model of representing uncertain information. It has extended the Dempster-Shafer theory \cite{Dempster1967,Shafer1976}. Dempster-Shafer theory is with advantages to handle uncertain information \cite{liu2006analyzing,schubert2011conflict,jousselme2012distances,yang2013evidential,Deng2014PAontheaxi,dubois2012conditioning,lefevre2013preserve,dezert2014validity}, and is extensively used in many fields, such as risk assessment \cite{zhang2012assessment,bolar2013condition}, expert systems \cite{yang2006belief,zhou2013bi}, classification and clustering \cite{denoeux2004evclus,bi2008combination}, parameter estimation \cite{denoeux2013maximum}, decision making \cite{casanovas2012fuzzy,mokhtari2012decision,yager2013decision}, and so forth \cite{srivastava2009representation,klein2012belief,cuzzolin2010geometry,cuzzolin2012relative,kang2012evidential,wei2013identifying,Deng2013TOPPER}. However, there are some weaknesses in Dempster-Shafer theory. D numbers overcome a few of existing deficiencies (i.e., exclusiveness hypothesis and completeness constraint) in Dempster-Shafer theory and appear to be more effective in representing various types of uncertainties. Some basic concepts about D numbers are given as follows.

\begin{definition}
Let $\Theta$ be a nonempty set $\Theta  = \{ F_1 ,F_2 , \cdots ,F_N \}$ satisfying $F_i \neq F_j$ if $i \neq j$, $\forall i,j = \{ 1, \cdots, N\}$ , a D number is a mapping formulated by
\begin{equation}
D: 2^{\Theta} \to [0,1]
\end{equation}
with
\begin{eqnarray}
\sum\limits_{B \subseteq \Theta } {D(B) \le 1}  \quad and \quad
D(\emptyset ) = 0
\end{eqnarray}
where $\emptyset$ is the empty set and $B$ is a subset of $\Theta$.
\end{definition}

If $\sum\limits_{B \subseteq \Theta } {D(B) = 1}$, the information expressed by the D number is said to be complete; if $\sum\limits_{B \subseteq \Theta } {D(B) < 1}$, the information is said to be incomplete. The degree of information's completeness in a D number is defined as below.

\begin{definition}
Let $D$ be a D number on a finite nonempty set $\Theta$, the degree of information's completeness in $D$ is quantified by
\begin{equation}
Q = \sum\limits_{B \subseteq \Theta } {D(B)}
\end{equation}
\end{definition}

For the sake of simplification, the degree of information's completeness of a D number is called as its $Q$ value.

For a discrete set $\Theta = \{b_1, b_2, \cdots, b_i, \cdots, b_n\}$, where $b_i \in R$ and $b_i \ne b_j$ if $i \ne j$, a special form of D numbers can be expressed by \cite{Deng2014DAHPSupplier,Deng2014EnvironmentDNs}
\begin{equation}
\begin{array}{l}
 D(\{ b_1 \} ) = v_1  \\
 D(\{ b_2 \} ) = v_2  \\
 \cdots \qquad \cdots \\
 D(\{ b_i \} ) = v_i  \\
 \cdots \qquad \cdots \\
 D(\{ b_n \} ) = v_n  \\
 \end{array}
\end{equation}
or simply denoted as $D = \{(b_1, v_1), (b_2, v_2), \cdots, (b_i, v_i), \cdots, (b_n, v_n) \}$, where $b_i \ne b_j$ if $i \ne j$, $v_i > 0$ and $\sum\limits_{i = 1}^n {v_i } \le 1$. Some properties of this form of D numbers are introduced as follows.

\begin{property}
Permutation invariability. If there are two D numbers that $D_1 = \{(b_1, v_1), \cdots, (b_i, v_i), \cdots, (b_n, v_n) \}$ and $D_2 = \{ (b_n, v_n), \cdots, (b_i, v_i), \cdots, (b_1, v_1)\}$, then $D_1 \Leftrightarrow D_2$.
\end{property}

\begin{property}
For a D number $D = \{(b_1, v_1), (b_2, v_2), \cdots, (b_i, v_i), \cdots, (b_n, v_n) \}$, the integration representation of $D$ is defined as
\begin{equation}\label{D_integration}
I(D) = \sum\limits_{i = 1}^n {b_i v_i }
\end{equation}
where $b_i \in R$, $v_i > 0$ and $\sum\limits_{i = 1}^n {v_i } \le 1$. For the sake of simplification, the integration representation of a D number is called as its $I$ value.
\end{property}

\subsection{D-CFPR: D numbers extended CFPR}
As mentioned above, the CFPR provides an option to establish the decision matrix which only requires $n-1$ pairwise comparisons. Moreover, the reciprocity and additive consistency have been preserved in a CFPR. However, the original CFPR is constructed on the foundation of complete and certain information. It is unable to deal with the cases involving incomplete and uncertain information. This deficiency also has existed in the fuzzy preference relation. For example, assume there are $n$ experts who were invited to evaluate alternatives $A_i$ and $A_j$. Consider these cases.

Case 1: $x$ experts evaluate that $A_i$ is preferred to $A_j$ with a degree $d_x$, the remainder $y$ experts evaluate that $A_i$ is preferred to $A_j$ with a degree $d_y$, where $x < n$ and $y = n - x$.

Case 2: $x$ experts evaluate that $A_i$ is preferred to $A_j$ with a degree $d_x$, the remainder $y$ experts do not give any evaluations due to the lack of knowledge, where $x < n$ and $y = n - x$.


Obviously, both the original CFPR and fuzzy preference relation are incapable of representing and handling the aforementioned cases. In \cite{Deng2014DAHPSupplier} we studies the deficiency in the situation of fuzzy preference relations, and proposed the concept of D numbers preference relations which extends the fuzzy preference relations by using D numbers in order to overcome this deficiency. In this paper, we concentrate on the deficiency in the situation of CFPRs. The D numbers extended CFPR, shorted for D-CFPR, is proposed to strengthen CFPR's ability of expressing uncertain information by using D numbers. the D-CFPR is formulated by
\begin{equation}\label{EqRDDCFPR}
R_D  = \begin{array}{*{20}c}
   {} & {\begin{array}{*{20}c}
   {A_1 \; } & {A_2 \; } &  \cdots  & {A_n }  \\
\end{array}}  \\
   {\begin{array}{*{20}c}
   {A_1 }  \\
   {A_2 }  \\
    \vdots   \\
   {A_n }  \\
\end{array}} & {\left[ {\begin{array}{*{20}c}
   {D_{11} } & {D_{12} } &  \cdots  & {D_{1n} }  \\
   {D_{21} } & {D_{22} } &  \cdots  & {D_{2n} }  \\
    \vdots  &  \vdots  &  \ddots  &  \vdots   \\
   {D_{n1} } & {D_{n2} } &  \cdots  & {D_{nn} }  \\
\end{array}} \right]}  \\
\end{array}
\end{equation}
where $D_{ij} = \{(b^1_{ij}, v^1_{ij}), (b^2_{ij}, v^2_{ij}), \cdots, (b^p_{ij}, v^p_{ij}), \cdots \}$, $D_{ji} = \neg D_{ij} = \{(1-b^1_{ij}, v^1_{ij}), (1-b^2_{ij}, v^2_{ij}), \cdots, (1-b^p_{ij}, v^p_{ij}), \cdots \}, \; \forall i,j \in \{ 1,2, \cdots ,n\}$, and $b_{ij}^p \in [0,1]$, $v_{ij}^p  > 0$, $\sum\limits_p {v_{ij}^p }  = 1$. Obviously, $D_{ii} = \{(0.5, 1.0)\} \;\; \forall i \in \{ 1,2, \cdots ,n\}$ in $R_D$.

In Eq.(\ref{EqRDDCFPR}), $R_D$ is called a D-CFPR because it is constructed based on $n - 1$ pairwise comparisons denoted as $\{D_{12}, D_{23}, \cdots, D_{(n-1)n}\}$ which is a set of D numbers. Here, a method  is proposed to implement the construction of D-CFPRs.

At first, for the elements in the set of $\{ D_{ji} ,1 \le i < j \le n\}$ in which $D_{ji}  = \{ (b_{ji}^1 ,v_{ji}^1 ),(b_{ji}^2 ,v_{ji}^2 ), \cdots ,(b_{ji}^k ,v_{ji}^k ), \cdots \}$, $D_{ji}$ is given by
\begin{equation}
D_{ji}  = \frac{{j - i + 1}}{2} \ominus D_{i(i + 1)}  \ominus D_{(i + 1)(i + 2)}  \ominus  \cdots  \ominus D_{(j - 1)j}
\end{equation}
in which every component $(b_{ji}^k, v_{ji}^k)$ is obtained by
\begin{equation}
b_{ji}^k  = \frac{{j - i + 1}}{2} - b_{i(i + 1)}^{x }  - b_{(i + 1)(i + 2)}^{y }  -  \cdots  - b_{(j - 1)j}^{z } \;, \quad \forall (x, y, \cdots, z)
\end{equation}
\begin{equation}
v_{ji}^k  = \sum\limits_{(x,y, \cdots ,z) \in \Omega } {v_{i(i + 1)}^x  \times v_{(i + 1)(i + 2)}^y  \times  \cdots  \times v_{(j - 1)j}^z }
\end{equation}
where $\Omega = \{ (x, y, \cdots, z) \; | \; b_{ji}^k  = \frac{{j - i + 1}}{2} - b_{i(i + 1)}^{x }  - b_{(i + 1)(i + 2)}^{y }  -  \cdots  - b_{(j - 1)j}^{z } \}$, and $(b_{i(i + 1)}^{x }, v_{i(i + 1)}^{x })$ is the $x$th component of $D_{i(i + 1)}$, $(b_{(i+1)(i + 2)}^{y }, v_{(i+1)(i + 2)}^{y })$ is the $y$th component of $D_{(i+1)(i + 2)}$, $\cdots \cdots$, $(b_{(j - 1)j}^{z }, v_{(j - 1)j}^{z })$ is the $z$th component of $D_{(j - 1)j}$.

At second, for the rest of entries in the D-CFPR, they can be calculated based on the reciprocal property, namely $D_{ji} = \neg D_{ij}$, $\forall i,j \in \{1,2,\cdots n\}$.

According to the aforementioned two steps, a D-CFPR can be constructed based on $n - 1$ pairwise comparisons $\{D_{12}, D_{23}, \cdots, D_{(n-1)n}\}$. For the generated D-CFPR, it is possible that some values of $b_{ij}$s  in these D numbers $\{D_{ij}, \;\; i,j \in \{1,2,\cdots, n\}\}$ do not fall in the interval $[0, 1]$, but fall in an interval $[-a, 1+a]$, $a>0$. In such a case, the values of all $b_{ij}$s in every D numbers need to be transformed by using a transformation function which is given in Eq.(\ref{Transfunction}). The transformation function works as normalization, which transforms the values of $b_{ij}$s from $[-a,1 + a]$ to the interval $[0, 1]$.

In summary, the proposed method to construct a D-CFPR $R_D$ on alternatives $\{A_1, A_2, \cdots, A_n, n \ge 2\}$ from $n-1$ preference values $\{D_{12}, D_{23}, \cdots, D_{(n-1)n}\}$ is implemented as the following steps:
\begin{enumerate}
  \item Start.  Let $\{D_{12}, D_{23}, \cdots, D_{(n-1)n}\}$ as input.
  \item Calculate the set of preference values $B$ as

  $B = \{ D_{ji} ,1 \le i < j \le n\}$,

  $D_{ji}  = \frac{{j - i + 1}}{2} \ominus D_{i(i + 1)}  \ominus D_{(i + 1)(i + 2)}  \ominus  \cdots  \ominus D_{(j - 1)j}$.
  \item $R_D^{'} = B \cup \neg B$. For $R_D^{'}$, if the values of all $b_{ij}$s in each D numbers fall in the interval $[0, 1]$, the D-CFPR $R_D$ is obtained as $R_D = R_D^{'}$, go to step 6; otherwise, go to next step.
  \item $a = \left| {\min \;\{ b_{ij} \;{\rm{in}}\;D_{ij} \} } \right|$, $D_{ij} \in R_D^{'}$, $i,j \in \{1,2,\cdots, n\}$.
  \item The D-CFPR $R_D$ is obtained as $R_D = f(R_D^{'})$ such that

  $f:[ - a,1 + a] \to [0,1]$,

  $f(b_{ij}) = \frac{{b_{ij} + a}}{{1 + 2a}}$, $\forall D_{ij} \in R_D^{'}$, $i,j \in \{1,2,\cdots, n\}$.
  \item End.
\end{enumerate}

Up to now, the method to construct a D-CFPR is totally presented. It should be pointed that the D-CFPR will reduce to the classical CFPR if the D numbers based preference values are substituted by real numbers. And the method to construct a D-CFPR is completely suitable for the construction of CFPRs. The proposed D-CFPR is an extension of the classical CFPR.

\subsection{Solution for the D-CFPR}
\begin{figure}[htbp]
\begin{center}
\psfig{file=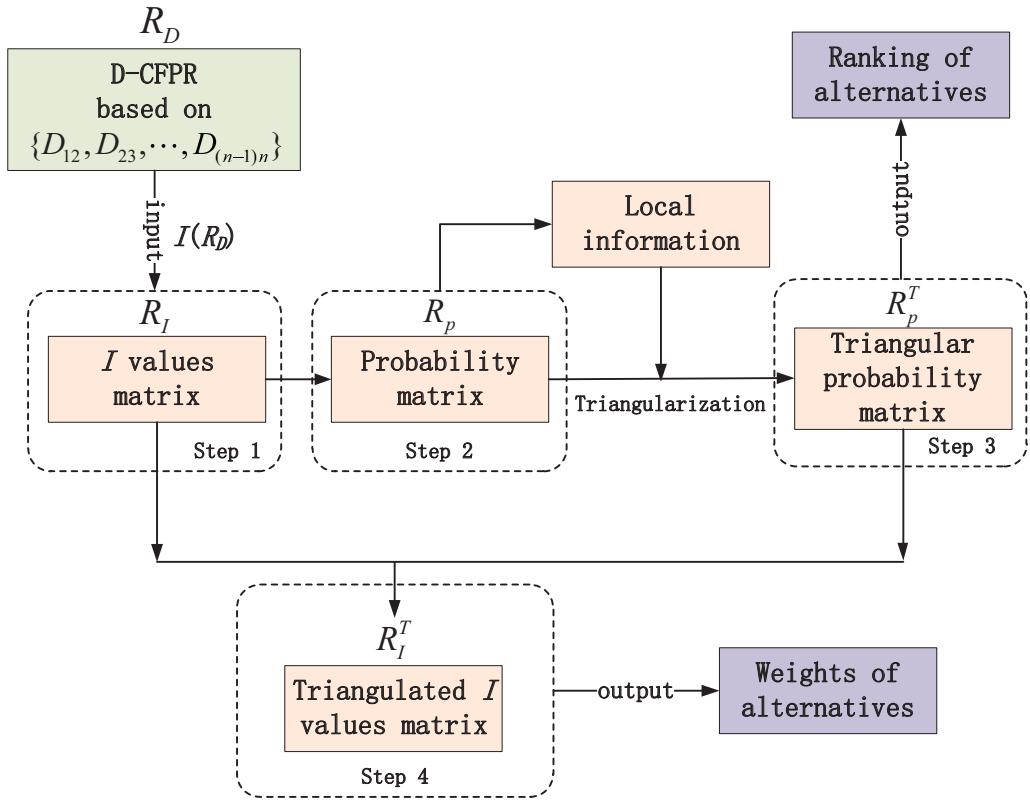,scale=0.6,angle=-90}
\caption{The procedure to obtain the ranking and priority weights of alternatives based on the D-CFPR \cite{Deng2014DAHPSupplier}}\label{FigSolutionForNCFPR}
\end{center}
\end{figure}

Once a D-CFPR has been constructed, another key problem is aroused that how to obtain the ranking and priority weights of alternatives based on the D-CFPR. In \cite{Deng2014DAHPSupplier}, we studied the solution for D numbers preference relations which extends the fuzzy preference relations by using D numbers. Differ from common D numbers preference relations, the D-CFPR is obtained based the additive transitivity property of CFPRs. The proposed D-CFPR in the paper essentially is an special case of D numbers preference relations. Therefore, the solution for D numbers preference relations is also suitable for D-CFPRs. The procedure of the solution for D-CFPRs is shown in Figure. \ref{FigSolutionForNCFPR}. At first, the D-CFPR $R_D$ is converted to an $I$ values matrix $R_I$ by using Eq.(\ref{D_integration}). At second, construct a probability matrix $R_p$ based on the $I$ values matrix $R_I$ to represent the preference probability between pairwise alternatives. At third, a triangular probability matrix $R_p^T$ can be obtained in terms of probability matrix $R_p$ with the aid of local information which contains the preference relation of pairwise alternatives. According to $R_p^T$, the ranking of alternatives is determined. At fourth, a triangulated $I$ values matrix $R_I^T$ is generated based on $R_I$ and $R_p^T$, and the weights of alternatives can calculated through $R_I^T$. Please refer to literature \cite{Deng2014DAHPSupplier} for more details.

\subsection{Inconsistency for the D-CFPR}
The classical CFPR is totally consistent due to it is constructed based on the additive transitivity property from $n - 1$ preference values. As an extension of CFPRs, the D-CFPR is also totally consistent when it has reduced to the classical CFPR. When the D-CFPR is with entries which contain uncertain or incomplete information, it is not totally consistent. In order to measure the inconsistency of D-CFPRs, an inconsistency degree $I.D.$ defined for D numbers preference relations \cite{Deng2014DAHPSupplier} is utilized to express such inconsistency. The inconsistency degree is on the basis of the triangular probability matrix $R_p^T$.
\begin{equation}
I.D. = \frac{{\sum\limits_{i = 1,j < i}^n {R_p^T (i,j)} }}{{n(n - 1)/2}}
\end{equation}

\section{Numerical examples}\label{SectNumeExams}
In this section, some numerical examples are given to show the construction of D-CFPRs.

\subsection{Example 1: preference values with certain information}
This example is from literature \cite{HerreraViedma2004someiss}. Suppose there is a set of four alternatives $\{A_1, A_2, A_3, A_4\}$ where we have certain knowledge to assure that alternative $A_1$ is weakly more important than alternative $A_2$, alternative $A_2$ is more important than $A_3$ and finally alternative $A_3$ is strongly more important than alternative $A_4$. Suppose that this situation is modelled by the preference values $\{r_{12} = 0.55, r_{23} = 0.65, r_{34} = 0.75\}$.

By applying the method proposed in \cite{HerreraViedma2004someiss}, a CFPR is obtained:
\begin{equation}\label{ExampleObtainedCFPR}
R = \left[ {\begin{array}{*{20}c}
   {0.5} & {0.55} & {0.7} & {0.95}  \\
   {0.45} & {0.5} & {0.65} & {0.9}  \\
   {0.3} & {0.35} & {0.5} & {0.75}  \\
   {0.05} & {0.1} & {0.25} & {0.5}  \\
\end{array}} \right]
\end{equation}

Howver, if the preference values are seen as a set of D numbers $\{D_{12} = \{(0.55, 1)\}, D_{23} = \{(0.65, 1)\}, D_{34} = \{(0.75, 1)\}\}$, a D-CFPR can be obtained by using the proposed method in this paper as follows.

$D_{31} = \{(0.3, 1)\}$ due to

$\qquad$ $b_{31}^1  = \frac{{3 - 1 + 1}}{2} - b_{12}^1  - b_{23}^1  = 1.5 - 0.55 - 0.65 = 0.3$,

$\qquad$ $v_{31}^1  = v_{12}^1  \times v_{23}^1  = 1 \times 1 = 1$.

$D_{41} = \{(0.05, 1)\}$ due to

$\qquad$ $b_{41}^1  = \frac{{4 - 1 + 1}}{2} - b_{12}^1  - b_{23}^1 - b_{34}^1 = 2 - 0.55 - 0.65 - 0.75 = 0.05$,

$\qquad$ $v_{41}^1  = v_{12}^1  \times v_{23}^1 \times v_{34}^1 = 1 \times 1 \times 1 = 1$.

$D_{42} = \{(0.1, 1)\}$ due to

$\qquad$ $b_{42}^1  = \frac{{4 - 2 + 1}}{2} - b_{23}^1  - b_{34}^1  = 1.5 - 0.65 - 0.75 = 0.1$,

$\qquad$ $v_{42}^1  = v_{23}^1  \times v_{34}^1  = 1 \times 1 = 1$.

$D_{21} = \neg D_{12} = \{(0.45, 1)\}$, $D_{13} = \neg D_{31} = \{(0.7, 1)\}$, $D_{14} = \neg D_{41} = \{(0.95, 1)\}$, $D_{32} = \neg D_{23} = \{(0.35, 1)\}$, $D_{24} = \neg D_{42} = \{(0.9, 1)\}$, $D_{43} = \neg D_{34} = \{(0.25, 1)\}$, and therefore:
\begin{equation}\label{ExampleObtainedDCFPR111}
R_D^{'} = \left[ {\begin{array}{*{20}c}
   {\{ (0.5,1)\} } & {\{ (0.55,1)\} } & {\{ (0.7,1)\} } & {\{ (0.95,1)\} }  \\
   {\{ (0.45,1)\} } & {\{ (0.5,1)\} } & {\{ (0.65,1)\} } & {\{ (0.9,1)\} }  \\
   {\{ (0.3,1)\} } & {\{ (0.35,1)\} } & {\{ (0.5,1)\} } & {\{ (0.75,1)\} }  \\
   {\{ (0.05,1)\} } & {\{ (0.1,1)\} } & {\{ (0.25,1)\} } & {\{ (0.5,1)\} }  \\
\end{array}} \right]
\end{equation}

For $R_D^{'}$, the values of all $b_{ij}$s in each D numbers fall in the interval $[0, 1]$, so the final D-CFPR $R_D = R_D^{'}$. D-CFPR $R_D$ is identical with CFPR $R$ shown in Eq.(\ref{ExampleObtainedCFPR}).

Finally, the method proposed in literature \cite{Deng2014DAHPSupplier} is utilized in order to obtain the priority weight of each alternative. The calculating process is omitted. Some important intermediate results are given as follows.
\begin{equation}
R_I  = \left[ {\begin{array}{*{20}c}
   {{\rm{0}}{\rm{.50}}} & {{\rm{0}}{\rm{.55}}} & {{\rm{0}}{\rm{.70}}} & {{\rm{0}}{\rm{.95}}}  \\
   {{\rm{0}}{\rm{.45}}} & {{\rm{0}}{\rm{.50}}} & {{\rm{0}}{\rm{.65}}} & {{\rm{0}}{\rm{.90}}}  \\
   {0.30} & {0.35} & {0.50} & {0.75}  \\
   {0.05} & {0.10} & {0.25} & {0.50}  \\
\end{array}} \right]
\end{equation}
\begin{equation}
R_p  = \left[ {\begin{array}{*{20}c}
   {\rm{0}} & {\rm{1}} & {\rm{1}} & {\rm{1}}  \\
   {\rm{0}} & {\rm{0}} & {\rm{1}} & {\rm{1}}  \\
   0 & 0 & 0 & 1  \\
   0 & 0 & 0 & 0  \\
\end{array}} \right]
\end{equation}
\begin{equation}
R_p^T  = \left[ {\begin{array}{*{20}c}
   {\rm{0}} & {\rm{1}} & {\rm{1}} & {\rm{1}}  \\
   {\rm{0}} & {\rm{0}} & {\rm{1}} & {\rm{1}}  \\
   0 & 0 & 0 & 1  \\
   0 & 0 & 0 & 0  \\
\end{array}} \right]
\end{equation}
\begin{equation}
R_I^T  = \left[ {\begin{array}{*{20}c}
   {{\rm{0}}{\rm{.50}}} & {{\rm{0}}{\rm{.55}}} & {{\rm{0}}{\rm{.70}}} & {{\rm{0}}{\rm{.95}}}  \\
   {{\rm{0}}{\rm{.45}}} & {{\rm{0}}{\rm{.50}}} & {{\rm{0}}{\rm{.65}}} & {{\rm{0}}{\rm{.90}}}  \\
   {0.30} & {0.35} & {0.50} & {0.75}  \\
   {0.05} & {0.10} & {0.25} & {0.50}  \\
\end{array}} \right]
\end{equation}

The inconsistency degree is $I.D. = \frac{{0.0}}{{4(4 - 1)/2}} = 0.0$. The obtained Priority weights and ranking of alternatives are shown in Table \ref{SolutionExampleONE}.

\begin{table}[htbp]
    \caption{Priority weights and ranking of alternatives in Example 1}\label{SolutionExampleONE}
    \begin{center}
    \begin{tabular}{cllllc}
    \toprule
    Alternatives & \multicolumn{4}{l}{Priority(different credibility of preference values)} & Ranking \\
    \cline{2-5}
    {} & High & Medium & Low & Interval range &  \\
    \midrule
    $A_1$ & 0.338 & 0.294 & 0.272 & (0.250, 0.409] & 1 \\
    $A_2$ & 0.312 & 0.281 & 0.266 & (0.250, 0.364] & 2 \\
    $A_3$ & 0.237 & 0.244 & 0.247 & [0.227, 0.250) & 3 \\
    $A_4$ & 0.112 & 0.181 & 0.216 & [0.000, 0.250) & 4 \\
    \bottomrule
    \end{tabular}
    \end{center}
\end{table}

\subsection{Example 2: preference values with uncertain and incomplete information}
In this example, we make a change to the preference values $\{r_{12} = 0.55, r_{23} = 0.65, r_{34} = 0.75\}$ from the above example so that the preference values are with uncertain and incomplete information. Assume this situation is modelled by a set of D numbers $\{D_{12} = \{(0.55, 0.8)\}, D_{23} = \{(0.65, 1)\}, D_{34} = \{(0.75, 0.9), (0.85, 0.1)\}$.

In this case, the preference values involve incomplete entry (i.e., $D_{12} = \{(0.55, 0.8)\}$) and uncertain entry (i.e., $D_{34} = \{(0.75, 0.9), (0.85, 0.1)\}$). The classical CFPR can not deal with this case, but the proposed D-CFPR is effective for this case. A D-CFPR can be obtained as follows.

$D_{31} = \{(0.3, 0.8)\}$ due to

$\qquad$ $b_{31}^1  = \frac{{3 - 1 + 1}}{2} - b_{12}^1  - b_{23}^1  = 1.5 - 0.55 - 0.65 = 0.3$,

$\qquad$ $v_{31}^1  = v_{12}^1  \times v_{23}^1  = 0.8 \times 1 = 0.8$;

$D_{41} = \{(0.05, 0.72), (-0.05, 0.08)\}$ due to

$\qquad$ $b_{41}^1  = \frac{{4 - 1 + 1}}{2} - b_{12}^1  - b_{23}^1 - b_{34}^1 = 2 - 0.55 - 0.65 - 0.75 = 0.05$,

$\qquad$ $v_{41}^1  = v_{12}^1  \times v_{23}^1 \times v_{34}^1 = 0.8 \times 1 \times 0.9 = 0.72$;

$\qquad$ $b_{41}^2  = \frac{{4 - 1 + 1}}{2} - b_{12}^1  - b_{23}^1 - b_{34}^2 = 2 - 0.55 - 0.65 - 0.85 = -0.05$,

$\qquad$ $v_{41}^2  = v_{12}^1  \times v_{23}^1 \times v_{34}^2 = 0.8 \times 1 \times 0.1 = 0.08$;

$D_{42} = \{(0.1, 0.9), (0.0, 0.1)\}$ due to

$\qquad$ $b_{42}^1  = \frac{{4 - 2 + 1}}{2} - b_{23}^1  - b_{34}^1  = 1.5 - 0.65 - 0.75 = 0.1$,

$\qquad$ $v_{42}^1  = v_{23}^1  \times v_{34}^1  = 1 \times 0.9 = 0.9$.

$\qquad$ $b_{42}^2  = \frac{{4 - 2 + 1}}{2} - b_{23}^1  - b_{34}^2  = 1.5 - 0.65 - 0.85 = 0.0$,

$\qquad$ $v_{42}^2  = v_{23}^1  \times v_{34}^2  = 1 \times 0.1 = 0.1$.

$D_{21} = \neg D_{12} = \{(0.45, 0.8)\}$, $D_{13} = \neg D_{31} = \{(0.7, 0.8)\}$, $D_{14} = \neg D_{41} = \{(0.95, 0.72), (1.05, 0.08)\}$, $D_{32} = \neg D_{23} = \{(0.35, 1)\}$, $D_{24} = \neg D_{42} = \{(0.9, 0.9), (1.0, 0.1)\}$, $D_{43} = \neg D_{34} = \{(0.25, 0.9), (0.15, 0.1)\}$, and therefore:

\begin{landscape}
\begin{equation}
R_D^{'}  = \left[ {\begin{array}{*{20}c}
   {\{ (0.5,1)\} } & {\{ (0.55,0.8)\} } & {\{ (0.7,0.8)\} } & {\{ (0.95,0.72),(1.05,0.08)\} }  \\
   {\{ (0.45,0.8)\} } & {\{ (0.5,1)\} } & {\{ (0.65,1)\} } & {\{ (0.9,0.9),(1.0,0.1)\} }  \\
   {\{ (0.3,0.8)\} } & {\{ (0.35,1)\} } & {\{ (0.5,1)\} } & {\{ (0.75,0.9),(0.85,0.1)\} }  \\
   {\{ (0.05,0.72),( - 0.05,0.08)\} } & {\{ (0.1,0.9),(0.0,0.1)\} } & {\{ (0.25,0.9),(0.15,0.1)\} } & {\{ (0.5,1)\} }  \\
\end{array}} \right]
\end{equation}

Due to the preference values in $R_D^{'}$ do not totally fall in the interval $[0, 1]$, but fall in an interval $[-0.05, 1.05]$, the transformation function shown in Eq.(\ref{Transfunction}) will be used to obtain the final D-CFPR $R_D$. The result is given as below.

\begin{equation}
R_D  = \left[ {\begin{array}{*{20}c}
   {\{ (0.5,1)\} } & {\{ (0.545,0.8)\} } & {\{ (0.682,0.8)\} } & {\{ (0.909,0.72),(1.0,0.08)\} }  \\
   {\{ (0.455,0.8)\} } & {\{ (0.5,1)\} } & {\{ (0.636,1)\} } & {\{ (0.864,0.9),(0.955,0.1)\} }  \\
   {\{ (0.318,0.8)\} } & {\{ (0.364,1)\} } & {\{ (0.5,1)\} } & {\{ (0.727,0.9),(0.818,0.1)\} }  \\
   {\{ (0.091,0.72),(0.0,0.08)\} } & {\{ (0.136,0.9),(0.045,0.1)\} } & {\{ (0.273,0.9),(0.182,0.1)\} } & {\{ (0.5,1)\} }  \\
\end{array}} \right]
\end{equation}
\end{landscape}

Similar with Example 1, in order to obtain the priority weight of each alternative, the method proposed in literature \cite{Deng2014DAHPSupplier} is employed. Some important intermediate results are given as follows.
\begin{equation}
R_I  = \left[ {\begin{array}{*{20}c}
   {{\rm{0}}{\rm{.5000}}} & {{\rm{0}}{\rm{.4360}}} & {{\rm{0}}{\rm{.5456}}} & {{\rm{0}}{\rm{.7345}}}  \\
   {{\rm{0}}{\rm{.3640}}} & {{\rm{0}}{\rm{.5000}}} & {{\rm{0}}{\rm{.6360}}} & {{\rm{0}}{\rm{.8731}}}  \\
   {0.2544} & {0.3640} & {0.5000} & {0.7361}  \\
   {0.0655} & {0.1269} & {0.2639} & {0.5000}  \\
\end{array}} \right]
\end{equation}
\begin{equation}
R_p  = \left[ {\begin{array}{*{20}c}
   {\rm{0}} & {{\rm{0}}{\rm{.68}}} & {\rm{1}} & {\rm{1}}  \\
   {{\rm{0}}{\rm{.32}}} & {\rm{0}} & {\rm{1}} & {\rm{1}}  \\
   0 & 0 & 0 & 1  \\
   0 & 0 & 0 & 0  \\
\end{array}} \right]
\end{equation}
\begin{equation}
R_p^T  = \left[ {\begin{array}{*{20}c}
   {\rm{0}} & {{\rm{0}}{\rm{.68}}} & {\rm{1}} & {\rm{1}}  \\
   {{\rm{0}}{\rm{.32}}} & {\rm{0}} & {\rm{1}} & {\rm{1}}  \\
   0 & 0 & 0 & 1  \\
   0 & 0 & 0 & 0  \\
\end{array}} \right]
\end{equation}
\begin{equation}
R_I^T  = \left[ {\begin{array}{*{20}c}
   {{\rm{0}}{\rm{.5000}}} & {{\rm{0}}{\rm{.5360}}} & {{\rm{0}}{\rm{.6456}}} & {{\rm{0}}{\rm{.8345}}}  \\
   {{\rm{0}}{\rm{.4640}}} & {{\rm{0}}{\rm{.5000}}} & {{\rm{0}}{\rm{.6360}}} & {{\rm{0}}{\rm{.8731}}}  \\
   {0.3544} & {0.3640} & {0.5000} & {0.7361}  \\
   {0.1655} & {0.1269} & {0.2639} & {0.5000}  \\
\end{array}} \right]
\end{equation}

In this example, the inconsistency degree is $I.D. = \frac{{0.32}}{{4(4 - 1)/2}} = 0.0533$. The obtained Priority weights and ranking of alternatives are shown in Table \ref{SolutionExampleTWO}.

\begin{table}[htbp]
    \caption{Priority weights and ranking of alternatives in Example 2}\label{SolutionExampleTWO}
    \begin{center}
    \begin{tabular}{cllllc}
    \toprule
    Alternatives & \multicolumn{4}{l}{Priority(different credibility of preference values)} & Ranking \\
    \cline{2-5}
    {} & High & Medium & Low & Interval range &  \\
    \midrule
    $A_1$ & 0.327 & 0.289 & 0.269 & (0.250, 0.402] & 1 \\
    $A_2$ & 0.309 & 0.280 & 0.265 & (0.250, 0.366] & 2 \\
    $A_3$ & 0.241 & 0.246 & 0.248 & [0.232, 0.250) & 3 \\
    $A_4$ & 0.123 & 0.186 & 0.218 & [0.000, 0.250) & 4 \\
    \bottomrule
    \end{tabular}
    \end{center}
\end{table}

\section{Concluding remarks}\label{SectConclusions}
In this paper, we have studied the consistent fuzzy preference relation (CFPR) by combining with D numbers. The proposed new preference relation is called D numbers extended consistent fuzzy preference relation, shorted for D-CFPR. A method is proposed to implement the construction of D-CFPR based on a set of preference values which are expressed by D numbers. In comparison with CFPR, D-CFPR is able to deal with the case that preference values involve incomplete and uncertain information. D-CFPR can be reduced to the classical CFPR when the preference values expressed by D numbers have degenerated to real numbers. What's more, based on our previous study in \cite{Deng2014DAHPSupplier}, the priority weights and ranking of alternatives can be obtained given a D-CFPR. In \cite{Deng2014DAHPSupplier}, the hierarchical structure of D-AHP model has been established. In this paper, the proposed D-CFPR can be employed to construct the preference matrix for D-AHP. Based on these two studies, the D-AHP model has been systematically built. In the future, we will focus on the application of proposed D-CFPR and D-AHP model.

\section*{Acknowledgements}
The work is partially supported by National Natural Science Foundation of China (Grant nos. 61174022 and 71271061), National High Technology Research and Development Program of China (863 Program) (Grant no. 2013AA013801), R \& D Program of China (2012BAH07B01).





\bibliographystyle{elsarticle-num}
\bibliography{references}







\end{document}